\documentclass[letterpaper, 10 pt, conference]{ieeeconf}  

\IEEEoverridecommandlockouts                              
\usepackage{graphicx}
\usepackage[]{algorithm2e}
\graphicspath{ {./} }

\title{\LARGE \bf
Value-of-Information based Arbitration between Model-based and Model-free Control
}

\author{Krishn Bera$^{1}$ Tejas Savalia$^{1}$ Bapi Raju$^{1*}$
\thanks{}
\thanks{$^{1}$ Cognitive Science Lab, IIIT-Hyderabad }%
\thanks{$^{*}$ Corresponding author: {\tt\small raju.bapi@iiit.ac.in}}%
}

\begin{document}

\maketitle
\thispagestyle{empty}
\pagestyle{empty}

\begin{abstract}

There have been numerous attempts in explaining the general learning behaviours using model-based and model-free methods. While the model-based control is flexible yet computationally expensive in planning, the model-free control is quick but inflexible. The model-based control is therefore immune from reward devaluation and contingency degradation. Multiple arbitration schemes have been suggested to achieve the data efficiency and computational efficiency of model-based and model-free control respectively. In this context, we propose a quantitative 'value of information' based arbitration between both the controllers in order to establish a general computational framework for skill learning. The interacting model-based and model-free reinforcement learning processes are arbitrated using an uncertainty-based value of information. We further show that our algorithm performs better than Q-learning as well as Q-learning with experience replay.

\end{abstract}

\section{INTRODUCTION}

Skill learning or skill acquisition is learning of a sequence of actions. A skill is learnt or improved when executed multiple number of times. An example of this could be playing a sport like long jump or bicycling. Since this could be seen as learning from a chain of events, it could be understood as a Markov Decision Process (MDP). It is formally defined as a 5-tuple $ [ S,A,P_{a},R_{a},\gamma] $ where $S$ is the finite set of possible states of the environment, $A$ is the finite set of actions available, $P_{a}(s,s')$ is the probability that action $a$ in state $s$ at time $t$ will lead to state $s'$ at time $t+1$, $R_{a}(s,s')$ is the immediate reward received after transitioning from state $s$ to state $s'$, due to action $a$ and $\gamma \in [0,1]$ is a discount factor which is used to make an infinite sum finite. In this context, the skill learning problem as a MDP can be solved using reinforcement learning (RL). Thus skill learning can be understood as a reinforcement learning problem where an agent needs to be trained to take a sequence of optimal decisions. 
\\
\\
In reinforcement learning, agents are trained by either model-based or model-free methods or a combination of both \cite{c1}\cite{c2}. Similarly, there have been many studies showing the role of two distinct learning processes in sensorimotor skill acquisition \cite{c3}\cite{c4}. In this paper we implement an algorithm that arbitrates between both model-based and model-free methods. The advantage of a model-based method is that it builds or assumes a model of environment dynamics (formally speaking, the agent has an estimate of $P_{a}$). Iterative methods like policy iteration or value iteration are used to solve MDPs in a model-based manner. The value-iteration exhaustively iterates through all the state and action spaces using a dynamic programming Bellman update equation to converge at an optimal value \cite{c5}. However, we may not know the environment dynamics in every situation and also it is computationally not feasible to iterate through all state-action when number of states increases dramatically (think Atari games). Hence, we use model-free methods where we sample trajectories and update the state-action values in every iteration using a temporal-difference learning equation \cite{c6}. Even though model-free methods are computationally inexpensive, they take a lot of time to converge i.e. it requires a substantial number of steps to explore and solve the environment. 
\\
\\
Previously there have been many attempts at arbitrating between the two processes in decision making, however, none of them categorically evaluated the role of dual processors in skill learning. For example, \cite{c7} represented the 'hybrid' learner's action value with weighted sum of action values of SARSA and FORWARD learner. \cite{c8} proposed an arbitration scheme that selects the dominant processor such that it outputs the optimal action with least uncertainty. The approach in \cite{c9} presents a 'plan-until-habit' strategy in order to balance the exploitation and exploration. In \cite{c10}, a value of information based arbitration is proposed in order to flexibly combine use of model-free(Q-learning) and model-based(sequential Monte-Carlo) methods for solving double T-maze environment.

\section{METHOD}
The basic idea behind our model is that we do a cost-benefit analysis of evaluating each action at a particular state to determine if model-based or model-free controller should be used for planning. Our interaction with the environment is set up in two stages: 'plan' and 'act'. The 'plan' stage is executed only when the agent is uncertain of it's actions given a state. The 'act' stage is a typical off-policy learning technique such as Q-Learning.The model-based method is implemented as a forward search technique using depth-limited search. When the depth is zero it acts like a normal off-policy algorithm and when depth is infinity it acts like a sampled value-iteration algorithm.

\subsection*{Value of Information (VoI) as a cost-benefit measure for planning}
In this paper we propose using VoI as a cost-benefit analysis parameter that helps the agent decide whether to execute model-based or model-free method at a given state. Since it is computationally expensive, we would like the agent to execute model-based method only when it is uncertain about taking a particular action. The model-free search is preferred when the agent is certain about the outcome of the particular action in the given state. Such cost-benefit-analysis is done and then the decision is made for every action at a given state. The arbitration is such that if VoI is above a threshold, it performs model-based search because the benefits outweigh the costs involved in executing it. A model-free planning is executed if VoI is below the threshold. The VoI is a directly proportional to variance of a state value given a state. The VoI is given by:
$$ VoI(s,a) = C_{(s,a)}/(\sigma(s)+\epsilon) $$
where $C_{(s,a)}$ is the uncertainty which is defined as the variance over the values of Q-values. The variance of the history of Q-values for a particular $(s,a)$ pair gives an estimate of uncertainty because if the trajectory involving that particular state-action pair is explored and estimated with a substantial confidence bound, the uncertainty (or variance) of the history of Q-values will be lesser than what would be the case otherwise if the exploration is not substantial and the estimation is uncertain. The quantity $\sigma(s)$ denotes the variance of the tuple $Q-values[s]$ for a particular state $s$. The value $(\sigma(s)+\epsilon)$ enables a systematic evaluation of a particular action $a$ in a given state $s$ if there is no clear and distinct candidate action in the tuple $Q-values[s]$, that can be chosen in order to maximize the agent's episodic reward.
\\
\\
In our planning stage we use value-iteration update but only considering those states within depth of two, which acts like a sampled value iteration every time planning is done.

\subsection*{Off-Policy learning for Acting}
We have used Q-learning as our model free control algorithm:
$$ Q(s,a) = Q(s,a) + \alpha * (r_{s,a} + max_{a}Q(s',a') - Q(s,a)) $$
where $Q(s,a)$ is the state-action value of the current state and the action taken and $Q(s',a')$ is the optimal next state-action value under the current policy.
\\
\\
In a nutshell, before an action is chosen (either by exploration or exploitation using a softmax function) we decide whether to update a state-action value by planning ahead. We hypothesize that model-based method is executed more often during initial episodes and decreases almost constantly with model free method being on the rise as number of episodes increases as shown in Fig. 1. This means that the environment is very uncertain initially and gets more familiar with more training. 

\subsection*{Environment}
For simulation purposes, we are using Taxi-v2 environment available in OpenAI gym \cite{c11}. It is a discrete, deterministic and fully observable environment with 500 states and 6 actions for each state. The episodic goal is to pick-up and drop-off the passenger to the right location while navigating the grid-world. Each time step results in -1 reward, each invalid pick-up or drop-off results in -10 reward and each successful episode gives +20 reward. 

\begin{algorithm}
Initialize Internal Model\;
\For{Iteration i}
{
  Reset the environment\;
  s = new state\;
  j = 0\;
  \While{Till reward reached}
  {
    \For{a in actions}
    {
        VoI(s,a) = $C_{(s,a)}/(\sigma(s)+\epsilon)$\;
        \eIf{ $VoI(s,a) \geq VoI_{threshold}$ }
        {
        	[Do model-based evaluation] \\
            Simulate policy execution using depth-limited search\;
            Use 'simulated reward evidence' to update transition probabilities and Q-table\;
        }
        {
        	[Do model-free evaluation] \\
            Retrieve the corresponding (cached) Q-value\;
        }
    }    

    Choose an action using softmax function on Q(s)\;
    Perform action\;
    Learn value of actions and states\;
    Update Q-table\;
  }
}
\caption{VoI based arbitration algorithm}
\end{algorithm}

\section{Results}
\subsection*{Arbitration between MB and MF controllers}
\begin{figure}[thpb]
\centering
\includegraphics[width=\linewidth]{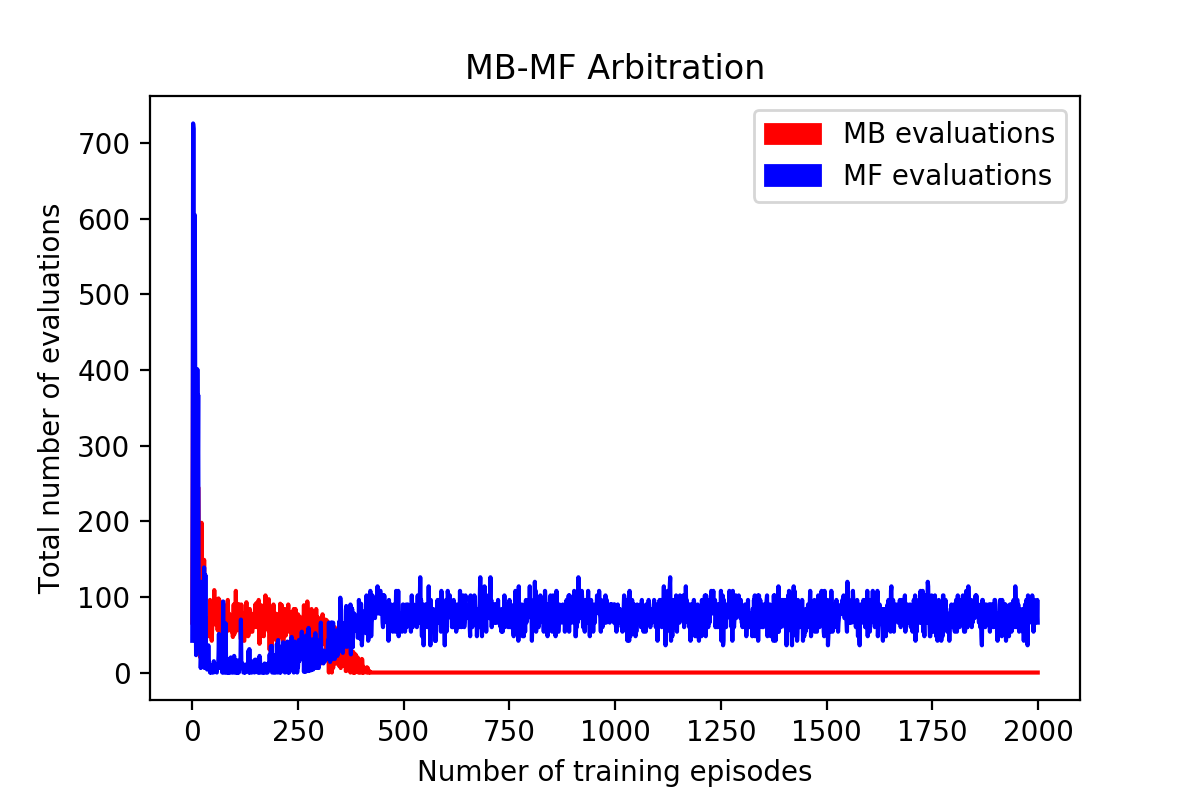}
\caption{MB-MF arbitration during the training}
\label{fig:mbmf_arbit}
\end{figure}

In the beginning, during the first training episode, almost equal number of model-based and model-free evaluations take place. Since the agent is just starting to learn the new environment without no previous experience, it realizes that model-based depth-limited-search will help in exploring the environment efficiently. The agent opts for the expensive but flexible model-based evaluations because not much is known about the environment dynamics. So initially, the VoI associated at each state-action pair will be high, thereby favouring model-based evaluations. We can see in Fig. \ref{fig:mbmf_arbit} that in the initial phase of the training, model-based evaluations dominate in comparison to a very few model-free evaluations.
\\

With further training, the agent has enough experience to learn the environment dynamics (the model of transition probabilities). The VoI now starts to decrease owing to the model-based exploration that the agent has already done in the initial phase. The algorithm now favours model-free exploration because it is fast, computationally inexpensive and also there is sufficient knowledge of the environment now. The model-based evaluations continuously decreases whereas the model-free evaluations increases with the training episodes. During the late phase of training, the environment is completely explored and known so doing expensive depth-limited-search would not be worth value agent achieves. Thus, model-based evaluations decrease to zero and remain so thereafter. The later phase of the training is dominated by model-free controller because it is relatively inexpensive in spite of the inflexible planning.

\subsection*{Performance Comparison}
Each experimentation consisted of averaging the total rewards obtained over 100 iterations of the complete episode of the Taxi environment. We ran 100 such experiments and averaged the rewards in order to measure the agent performance. The corresponding standard deviation was measured and plotted. We further do a side-by-side analysis of Q-learning agent as well as Q-learning with experience replay agent. 

\begin{figure}[thpb]
\centering
\includegraphics[width=\linewidth]{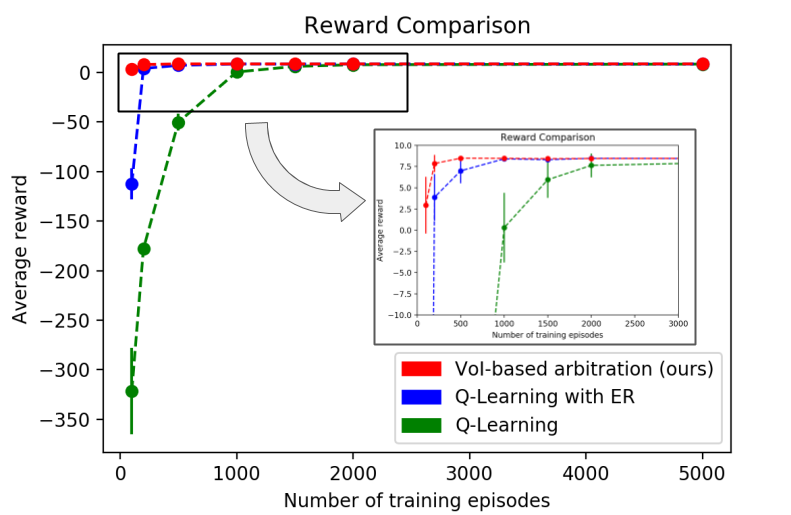}
\caption{Average reward received during training}
\label{fig:reward_in_img}
\end{figure}

As evident from Fig. \ref{fig:reward_in_img} (see Appendix for in-image), our VoI-based arbitration algorithm was the fastest to converge and solve the environment. While the naive Q-learning agent performed poorly over 500 training episodes with an average reward of -50.31$\pm$8.71, our algorithm received an average reward of 8.49$\pm$0.25. This is a significant difference in terms of learning and performance. Even though the Q-learning with ER agent outperformed the naive Q-learning agent with an average reward of 7.00$\pm$1.48 over 500 training episodes, it was not able to match the performance of VoI-based arbitration algorithm. Even from a convergence perspective, our algorithm required less than 500 training episodes whereas Q-learning with experience replay requires almost 1000 episodes to learn the environment. The Q-learning algorithm is the slowest to converge requiring almost 2500 episodes to converge.

\section{Discussion}
In the typical discrete skill learning task, the stimuli-response(S-R) outputs are slow at first due to the cognitive phase of execution where the cognitive processor is dominant. The average response time in the initial training phase is reasonably high. This is mostly attributed to the model-based depth-limited search. The results shown are for max depth set to 2. Increasing the depth will make the agent learn the environment dynamics quickly but it incurs more computational cost. From a cognitive perspective, we can consider this limited-depth search as the limitation of our cognitive resources (more specifically working memory). The model-based steps routinely take more time because they are data-efficient especially when there is not much information available about the environment at the start of the experiment. However, a purely model-based learning will follow the minimal number of steps to reach the goal state provided the optimal actions it takes by calculating the state prediction error. The model-free is quick and inexpensive search and is more suited to the later phase of the training when there is enough experience of the environment. However, in the later phase of the training, the execution speed gradually increases to attain a maximum after substantial practice. This is because the actions are now executing in motor mode which is autonomously controlled by the motor processor. We see a minimal cognitive processor activity in this phase as the motor processor executes a sequential learning task in an open-looped control. The motor processing takes a significantly lesser execution time at each step because it uses experience directly in form of reward prediction error. This is indicative of the fact that the model-free process has taken over from the earlier model-based reinforcement learning process. Therefore, the model-based reinforcement learning can be taken to obtain a typical representation of the early phase practice in sequential skill learning task such as Discrete Sequence Production.

\begin{figure}[thpb]
\centering
\includegraphics[width=\linewidth]{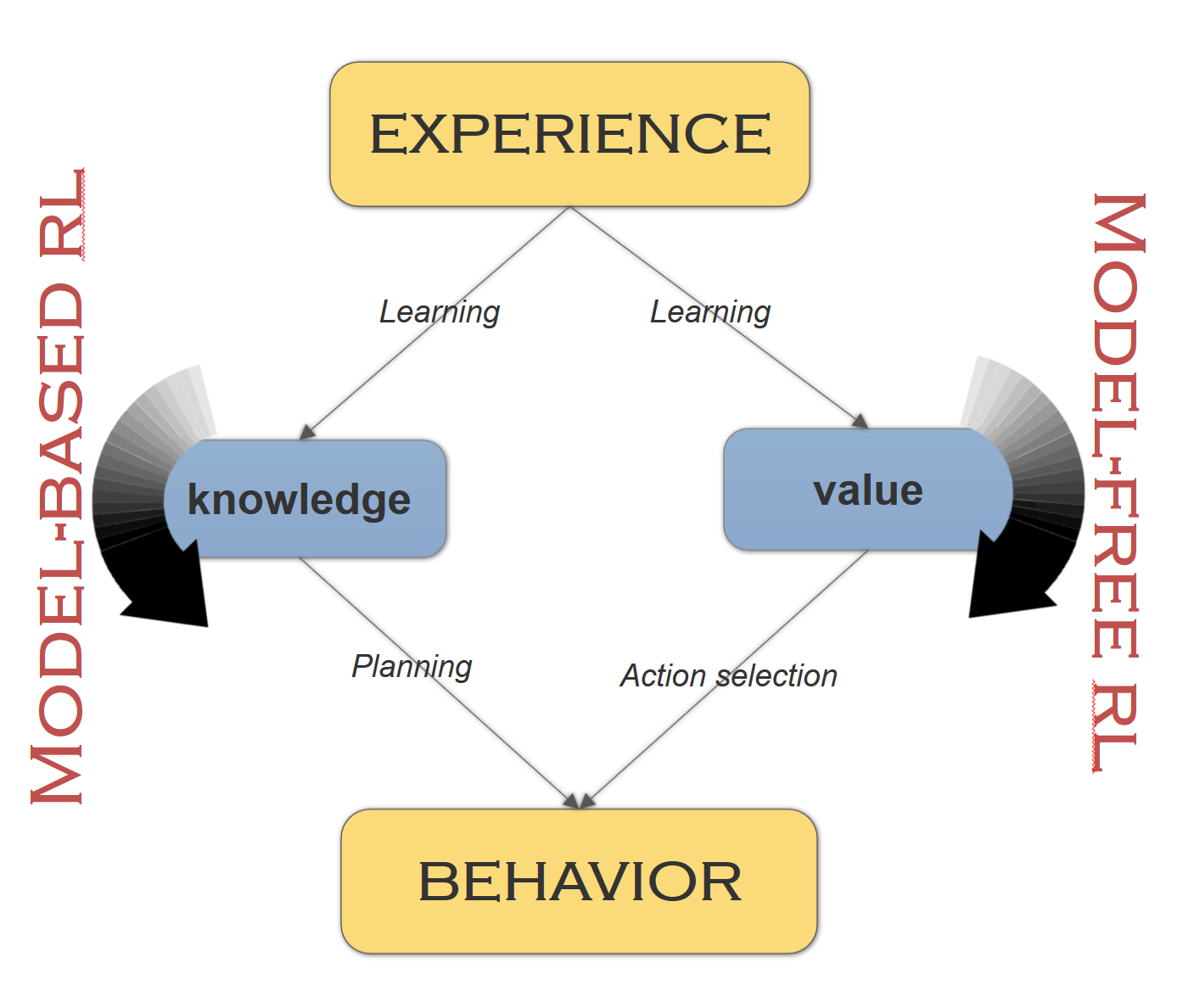}
\caption{Model-free and model-based Reinforcement Learning. Ref. Toussiant (2010)}
\label{fig:overview}
\end{figure}

The VoI estimate improves as the learning progresses. Learning the 'world-model' implies learning the transition probabilities table (in terms of MDP formulation). In the initial phase of the training when the model-based learning dominates, our framework is also capable of learning the transition probabilities. However, in case of OpenAI-Taxi environment, since the world-model is deterministic, it really doesn't matter if we are learning transitions as well. Thus, our framework is able to learn transitions alongside Q-values in order to provide a reliable estimate of good-ness of taking a particular action in a given state.
\\
\\
The model-based reinforcement is flexible and immune to any changes in the environment (say, reward devaluation or contingency degradation). It will accordingly try to adapt the action strategies. If we change the reward that the agent is getting on taking a particular action in some state, the q-values will reflect the change and thereby the value of VoI will adjust accordingly to favour model-based learning in order to update the model of the environment. 
\\
\\
Our model agrees with the typical behavioural phenomenon in Fitts' three phases of learning \cite{c12}. The initial phase is cognitive phase which is dominated by model-based learning processes. The planning during this phase is data-efficient and adaptive with a substantial computing cost. The intermediate associative phase is where we see a change of dominant processor - from model-based to model-free planner. The later motor phase is dominated by model-free processor. The planning during this phase is inexpensive but not flexible. As seen from Fig. \ref{fig:mbmf_arbit}, the model-based processor is dominant for almost 200-300 training episodes, after which the control is slowly arbitrated to the model-free planner. After about 400 episodes, model-based evaluations reach zero and all the behaviour is guided by model-free processor. The model-based evaluations reaching zero indicate the transition of arbitration mechanism from goal-oriented to habitual. It implies that the agent has already learned the environment and the reward pay-offs for taking a particular action in a given state and therefore, it would be an optimal strategy (in terms of speed-accuracy trade-off) to employ computationally inexpensive model-free reinforcement. Behaviourally, it is analogous to the last phase of the proposed Fitts’ theory of skill learning.
\\
\\
Comparing our framework with the others from the literature, we can argue that VoI-based arbitration is a more biologically-plausible framework of dual-processor model. Also, it is capable of validating many testable prediction from previous studies. For example, the Daw task proposed in \cite{c7} hypothesized and argued for the existence of two learning strategies involved in guiding our learning behaviour. However, it is not clear what kind of learning strategies are employed during various stages of learning. Our hybrid model is able to predict that the early phase of training is dominated by goal-oriented model-based processes whereas the later phase is dominated by habitual model-free processes.

\section{Conclusion}
In conclusion, our algorithm tries to balance the search efficiency of model-based planning and computational efficiency of model-free planning. The scope of future work can include proofs of convergence in order to establish that the arbitration happens properly. This arbitration is controlled by VoI measure that dynamically changes depending upon the variance of Q-value history and the Q-value difference among different actions available in a particular state. Such a scheme provides a systematic combining model-based and model-free mechanisms in one algorithmic model. Moreover, the results show that the performance of the agent agrees with the typical behavioural phenomenon of skill learning as shown in the literature.

\addtolength{\textheight}{-12cm}   



\section*{Appendix}

\begin{figure}[htb]
  \includegraphics[width=\linewidth]{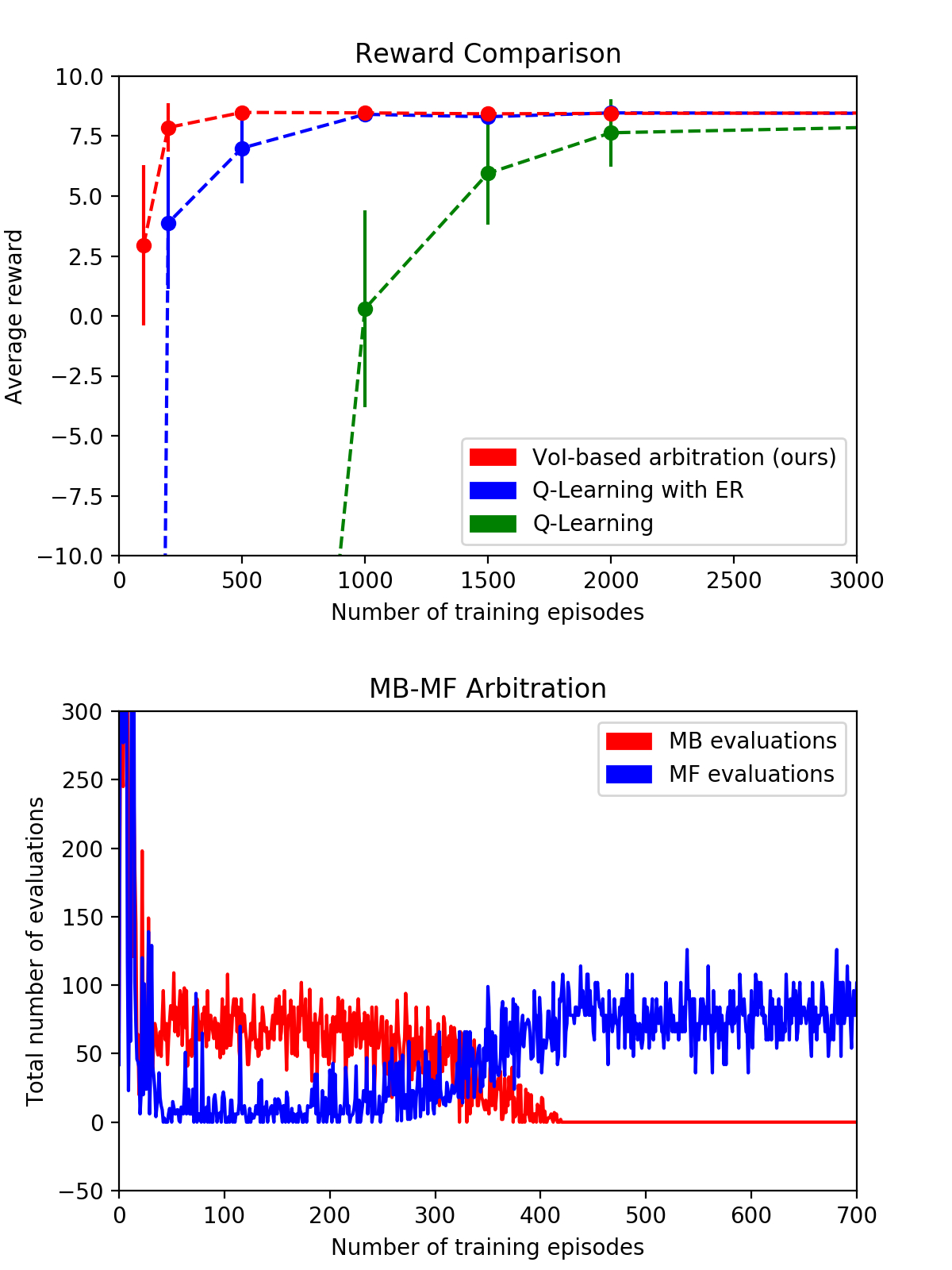}
  \caption{Results (enhanced)}
\end{figure}

\begin{table}[htb]
\centering
\caption{\bf List of Parameters}
    \begin{tabular}{ccc}
    \hline
    Hyper-parameter & Notation & Value \\
    \hline
     Learning rate & $\alpha$ & 0.8 \\ 
     Discount rate & $\gamma$ & 0.9 \\ 
     Max. depth & $d$ & 2 \\ 
     Inverse temperature & $\rho$ & 0.9 \\ 
     VoI threshold & $VoI_{threshold}$ & 0.1 \\ 
     VoI threshold multiplier & $VoI_{mult}$ & 1.005 \\
     Epsilon & $\epsilon$ & 0.000001 \\ [1ex] 
    \hline
    \end{tabular}
\end{table}

\end{document}